\documentclass[lettersize,journal]{IEEEtran}
\usepackage{amsmath,amsfonts}
\usepackage{algorithmic}
\usepackage{algorithm}
\usepackage{array}
\usepackage[caption=false,font=normalsize,labelfont=sf,textfont=sf]{subfig}
\usepackage{textcomp}
\usepackage{stfloats}
\usepackage{url}
\usepackage{verbatim}
\usepackage{graphicx}
\usepackage{cite}
\usepackage{booktabs,multirow}
\begin{document}

\title{Label-Guided Knowledge Distillation for 3D-CNNs in Action Recognition}

\author{IEEE Publication Technology,~\IEEEmembership{Staff,~IEEE,}

\author{
	Yanjiang~Shi,
	Peng~Zhao~\IEEEmembership{Member,~IEEE,}
    Nan~Qi,
    Guiqin~Wang,
	\thanks{Y. Shi is with the School of Computer Science and Technology, and with the National Engineering Laboratory for Big Data Analytics (NEL-BDA), Xi'an Jiaotong University, Xi'an 710049, China (Email: syj2908@stu.xjtu.edu.cn).}
     \thanks{Nan Qi and G. Wang are with the School of Computer Science and Technology, and with the National Engineering Laboratory for Big Data Analytics (NEL-BDA), Xi'an Jiaotong University, Xi'an 710049, China (Email: qinan@stu.xjtu.edu.cn; gqwang@stu.xjtu.edu.cn).}
     \thanks{P. Zhao is with the School of Computer Science and Technology, and with the National Engineering Laboratory for Big Data Analytics (NEL-BDA), Xi'an Jiaotong University, Xi'an 710049, China (Email: p.zhao@mail.xjtu.edu.cn).}
}

\thanks{Manuscript received April 19, 2021; revised August 16, 2021.}}

\markboth{Journal of \LaTeX\ Class Files,~Vol.~14, No.~8, August~2021}%
{Shell \MakeLowercase{\textit{et al.}}: A Sample Article Using IEEEtran.cls for IEEE Journals}


\maketitle

\begin{abstract}
As a key model compression technique, knowledge distillation aims to transfer knowledge from a high-capacity teacher model to a lightweight student model for enhancing the latter's performance. In this work, we reviewed the feature knowledge distillation for 3D-CNNs and observed that most feature distillation methods in video analysis are simple adaptations of those used in image analysis, often neglecting the differences of video features in the temporal dimension. To address this issue, we proposed Label-Guided Knowledge Distillation (LGKD) to guide the distillation of student model features using ground truth labels. Our method entails two components: sample-wise distillation and class-wise distillation, enabling the student model to learn feature representation of the teacher model at two levels. Sample-wise distillation utilizes label information and the teacher’s probability distribution to guide the learning of features that significantly impact temporal accuracy while mitigating noise. Meanwhile, class-wise feature distillation employs a prototype network to further capture the relational knowledge among samples within the same category, enhancing the student's ability to learn higher-dimensional semantic information and improving model generalization. To demonstrate the effectiveness and superiority of our method, we conducted comprehensive experiments on two benchmark action recognition datasets, UCF101 and HMDB51, achieving competitive results. 
\end{abstract}

\begin{IEEEkeywords}
Knowledge Distillation, Model Compression, Prototype Network
\end{IEEEkeywords}

\section{Introduction}
\IEEEPARstart{I}{n} recent years, deep learning has been increasingly applied to video understanding. As a fundamental task in this field, action recognition has garnered significant attention from both academia and industry. Deep learning models for video analysis can be broadly categorized into two architectures: 3D convolutional neural networks (3D-CNNs) and video transformers~\cite{zhao2024videoprism}. While 3D-CNNs have evolved considerably, achieving higher accuracy through increased model depth, this also results in a greater number of parameters and higher computational complexity. Consequently, deploying these high-precision models on resource-constrained devices remains challenging. Knowledge distillation (KD), a key model compression technique, addresses this issue by transferring knowledge from a high-accuracy, computationally intensive teacher model to a lightweight, lower-accuracy student model, thereby improving the latter’s performance. As a result, knowledge distillation for 3D-CNNs continues to be a valuable research direction.

Feature-based knowledge distillation has gained prominence recently as an effective approach due to the rich learnable information contained in feature maps~\cite{romero2014fitnets, ji2021refine,chung2020feature}. This method enables the student model to learn the teacher model’s ability to capture data representations by mimicking its feature map distribution. Current research on feature-based knowledge distillation primarily focuses on the image domain, particularly on model compression for 2D-CNNs. In contrast, knowledge distillation for 3D-CNNs remains relatively underexplored, with recent studies shifting toward cross-modal knowledge distillation. This approach utilizes feature maps as intermediaries to transfer knowledge from the optical flow branch to the RGB branch, making it applicable only to teacher-student networks designed for multi-modal data.

In the standard distillation paradigm, where knowledge is transferred from a larger teacher model to a smaller student model, existing methods largely adapt feature-based knowledge distillation techniques developed for 2D-CNNs. However, unlike static images, video data include a temporal dimension, which affects model inference in tasks such as action recognition. The importance of different frames within a video varies; for instance, in the action “cliff diving,” preparatory movements, takeoff, and descent contribute differently to classification. This variation, as observed in \cite{shao2023action}, suggests that labeled video frames can be categorized based on their significance for classification versus regression. The accuracy of features extracted from the latter has a greater impact on the final classification outcome. Despite this, conventional feature-based knowledge distillation methods treat all features extracted by the teacher model equally, potentially introducing noise during knowledge transfer. Motivated by this phenomenon, we aim to investigate how adjusting distillation loss weights based on the importance of video frame features can enhance knowledge distillation.

To address the aforementioned problem, we began with a simple validation experiment by incorporating label information into the distillation process. Unlike conventional methods that use label information solely to calculate the classification loss of the student model, we conduct knowledge distillation only on samples where the teacher model makes correct predictions. Specifically, ground truth labels are used to select the samples for distillation, with results presented in Table~\ref{tab1}. It can be observed that simply introducing label information enhances the effectiveness of both response-based and feature-based knowledge distillation. Notably, by leveraging ground truth labels and distilling only on correctly classified samples, accuracy improves by approximately 0.4\% in both distillation methods. This observation motivates us to further utilize label information to help the student model differentiate the importance of various frame features.

\begin{table}[htbp]\centering
\caption{The validation results on the UCF101 dataset.}
\label{tab1}
    \setlength{\tabcolsep}{1.2mm}{
    \begin{tabular}{ccc}
    \toprule
    \textit{Method} & \textit{Top-1}(\%) & \textit{Top-5}(\%)\\
    \midrule
    Baseline Student & 63.36 & 82.08\\
    Logits-based KD & 67.20 & 85.44\\
    Logits-based KD w/o Wrong Prediction & \textbf{67.57} & \textbf{86.18}\\
    Feature-based KD & 67.72 & 86.47\\
    Feature-based KD w/o Wrong Prediction & \textbf{68.20} & \textbf{86.60}\\
    \bottomrule
\end{tabular}}
\end{table}

\begin{figure}[!t]
    \centering
    \includegraphics[width=0.44\textwidth]{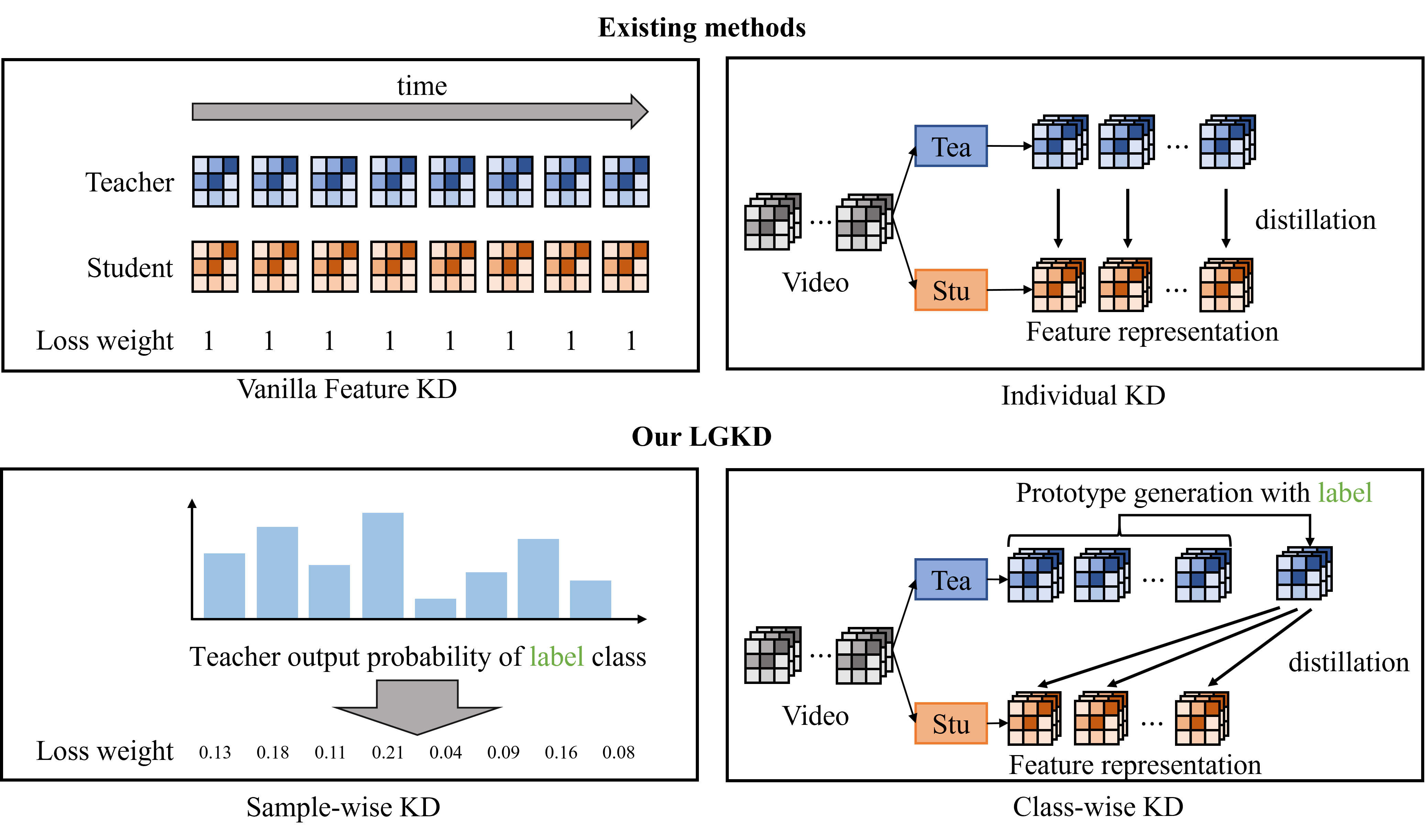}
    \caption{Label-guided knowledge distillation method. Compared to traditional feature distillation, sample-wise distillation leverages label information and the teacher's probability distribution to explore the temporal importance differences in teacher features. Meanwhile, class-wise distillation, rather than distilling knowledge independently for each sample, leverages label information to generate class prototypes that represent high-level semantic features.}
    \label{fig:intro 1}
\end{figure}

Motivated by this observation, we propose Label-Guided Knowledge Distillation(LGKD), a method tailored for action recognition tasks. As illustrated in Fig.~\ref{fig:intro 1}, LGKD performs feature distillation at both the sample and class levels, leveraging label information to fully extract learnable information from the teacher's features. Specifically, in sample-wise feature distillation, based on traditional feature distillation, we use label information to extract the probability distribution of the ground truth category inferred by the teacher model along the temporal dimension. This distribution serves as the basis for calculating importance scores of frame features, which are then used to assign corresponding weights to the feature loss. Additionally, to prevent the student model from overfitting the teacher's features during the sample-wise distillation, we introduce class-wise distillation loss. Considering samples within the same category have similarities, aligning features at the category level can guide the student model to capture high-order semantic information, making the feature representations of each category more compact while improving the generalization of the student model\cite{yun2020regularizing}. To achieve this, we propose a prototype network that transfers relational knowledge within each category. By filtering samples correctly classified by the teacher using ground truth labels, we generate action category feature prototypes—or category feature centers—to guide the student model. Additionally, we introduce a weight adjustment module in sample-wise feature distillation to dynamically adjust temporal loss weights for each teacher-student feature pair. In class-wise feature distillation, the prototype network generates prototype features for each action category, further enhancing knowledge transfer.

To the best of our knowledge, we are the first to consider temporal weighting in feature distillation and introduce label information to the distillation for video understanding tasks. Our main contributions are summarized as follows:

\begin{itemize}
\item To extract frame features that are more critical for classification during knowledge distillation, we propose a sample-wise knowledge distillation approach and design a weight adjustment module. This module leverages label information and the teacher’s probability distribution to guide the student model in learning features that are more important for classification accuracy in the temporal dimension.
\item We propose a class-wise knowledge distillation method based on a prototype network to mitigate feature divergence caused by sample-wise distillation. By leveraging label information to filter the features of samples correctly inferred by the teacher. This approach aligns features at the category level, facilitates the transfer of relational knowledge among samples within the same category, and enables the student model to learn higher-order semantic information from the teacher model.
\item We conduct extensive experiments on two widely used action recognition datasets, UCF101 and HMDB51. The experiment results demonstrate that our method significantly improves the feature representation capability of the student model, thereby surpassing the performance of other compared methods. Furthermore, we additional explore experiments to validate the effectiveness of our approach and its generalization capability across different teacher-student model pairs.
\end{itemize}

\section{Related Works}
\subsection{Action Recognition}
As a fundamental task in video understanding, action recognition has recently transitioned from purely 3D-CNN-based models to Transformer-based models\cite{wang2024clip}. However, the latter still require 3D-CNNs to extract accurate features\cite{liu2023survey}. The feature extraction networks of existing 3D-CNN architectures can be divided into fully 3D-CNNs\cite{carreira2017quo,xu2017r,ji20123d,hara2017learning} and partially 3D-NNs\cite{tran2018closer,qiu2017learning,xie2018rethinking}. Fully 3D-CNNs treat video as a sequence of continuous frames and use 3D convolutional kernels to extract representation, with typical models like I3D\cite{carreira2017quo}, R-C3D\cite{xu2017r}. Partially 3D-CNNs replace 3D convolutional layers with 2D convolutional layers to reduce computation costs and parameter requirements, i.e., Top-I3D, Bottom-I3D, and S3D\cite{xie2018rethinking}. However, while reducing resource demands, these partial 3D-CNNs also suffer from a noticeable decline in accuracy, necessitating further exploration to narrow the gap between fully 3D-CNNs and partially 3D-CNNs.
\subsection{Knowledge Distillation}
Knowledge distillation was first proposed by Hinton \textit{et al.}. As a branch of model compression technology, its purpose is to transfer the knowledge contained in the heavyweight, high-precision teacher model to the lightweight, low-precision student model to improve the student's accuracy\cite{hinton2015distilling}. Based on the form of the knowledge, knowledge distillation can be mainly categorized into response-based knowledge distillation, feature-based knowledge distillation, and api-based knowledge distillation\cite{tang2024survey}. Api-based knowledge distillation has emerged with the development of Large Language Models(LLMs), where the LLM acts as a black-box teacher to guide the student model. However, this method is only applicable to the teacher-student pairs which both of them are LLM\cite{gu2024minillm,xu2024survey}. Response-based knowledge distillation involves having the student model mimic the probability distribution outputted by the teacher to learn the "dark knowledge" contained in soft labels\cite{hinton2015distilling,li2023curriculum, sun2024logit}. In recent years, feature-based knowledge distillation has become the mainstream research direction due to its inclusion of more learnable information\cite{romero2014fitnets}. It uses feature maps as the medium of knowledge, enabling the student model to learn the teacher model's feature extraction capabilities by mimicking the distribution of the teacher model's feature maps.

In the research of feature distillation for convolutional neural networks, most current work still focuses on image data\cite{xu2017r,yang2022masked,chen2022knowledge}. In the field of video analysis, the mainstream research method is cross-modal knowledge distillation\cite{crasto2019mars,stroud2020d3d,lee2023decomposed,dai2021learning}. These methods treat the model branch that processes optical flow data as the teacher, aiming to transfer its feature extraction capabilities to the model branch that handles RGB data, thereby improving the accuracy of the unimodal model. Moreover, most feature distillation methods aimed at model compression simply adapt 2D-CNN distillation techniques, neglecting the differences in the importance of features across the temporal dimension\cite{ullah20233dcnn}.
\subsection{Prototype Network}
Prototype network was first applied to few-shot classification\cite{snell2017prototypical}, where it computes the distance between new samples and every class prototype in a learnable metric space for classification. Due to its simplicity and strong generalization capability, the prototype network has been extended to various tasks, such as transfer learning\cite{zhang2021prototypical} and zero-shot learning\cite{wang2022prototype}. ProDA\cite{zhang2021prototypical} introduces the concept of prototype network into pseudo-label denoising. PSKD\cite{wang2022prototype} utilizes prototype network to refine the teacher's probability distribution through a correlation matrix in zero-shot learning tasks. In the field of knowledge distillation, ProKD\cite{ge2023prokd} proposes a contrastive learning-based prototype alignment method to enhance the student model's ability to acquire specific language knowledge in zero-resource named entity recognition task. In our work, we utilize the prototype network to extract relational knowledge among samples within each action category, guiding the student model to learn the teacher model's feature representation capabilities.

\section{Methodology}

In this section, we first review the basic concepts of feature knowledge distillation and then introduce our label-guided knowledge distillation.
\subsection{Vanilla Feature Knowledge Distillation}
As a key method in model compression technology, knowledge distillation has been widely applied to many visual tasks. As a branch of knowledge distillation, feature knowledge distillation has increasingly become the mainstream method due to its excellent accuracy improvement. The primary objective of feature knowledge distillation is to transfer the knowledge contained in the heavyweight, high-precision teacher model to the lightweight, low-precision student model via feature maps\cite{heo2019comprehensive}. 

Formally, let $T$ and $S$ represent the teacher and student models, respectively. When video data $i$ is sent to the models, they generate feature maps $F^T$ and $F^S$, where $F\in R^{T\times C\times H\times W}$. Here, $T$,$C$,$H$ and $W$ represent the temporal, channel, and spatial dimensions, respectively. In traditional feature knowledge distillation, the distillation loss is composed of the $L_2$ distance between the feature maps produced by the teacher and student models:
\begin{equation}
\label{equ:vanilla-feat-kd}
    \mathcal{L}_{vanilla\ feature\ KD} = \sum_{n=1}^{N}\Vert f(F^T_{i,n})-f(F^S_{i,n})\Vert _2^2,
\end{equation}
where $f\left(\cdot\right)$ is a projection function that aligns the feature maps of the teacher and student models to the same dimension, $N$ represents the number of frames contained in the video. This constraint encourages the student model to mimic the feature map distribution of the teacher model, thereby capturing the teacher model's feature extraction capability. However, vanilla feature distillation method originally designed for image data applies equal weight to all feature maps during distillation. When transferred to video data domain that includes temporal dimension, it overlooks the fact that different video frame features may have different importance to the inference result.

\begin{figure*}[!t]
    \centering
    \includegraphics[width=0.88\textwidth]{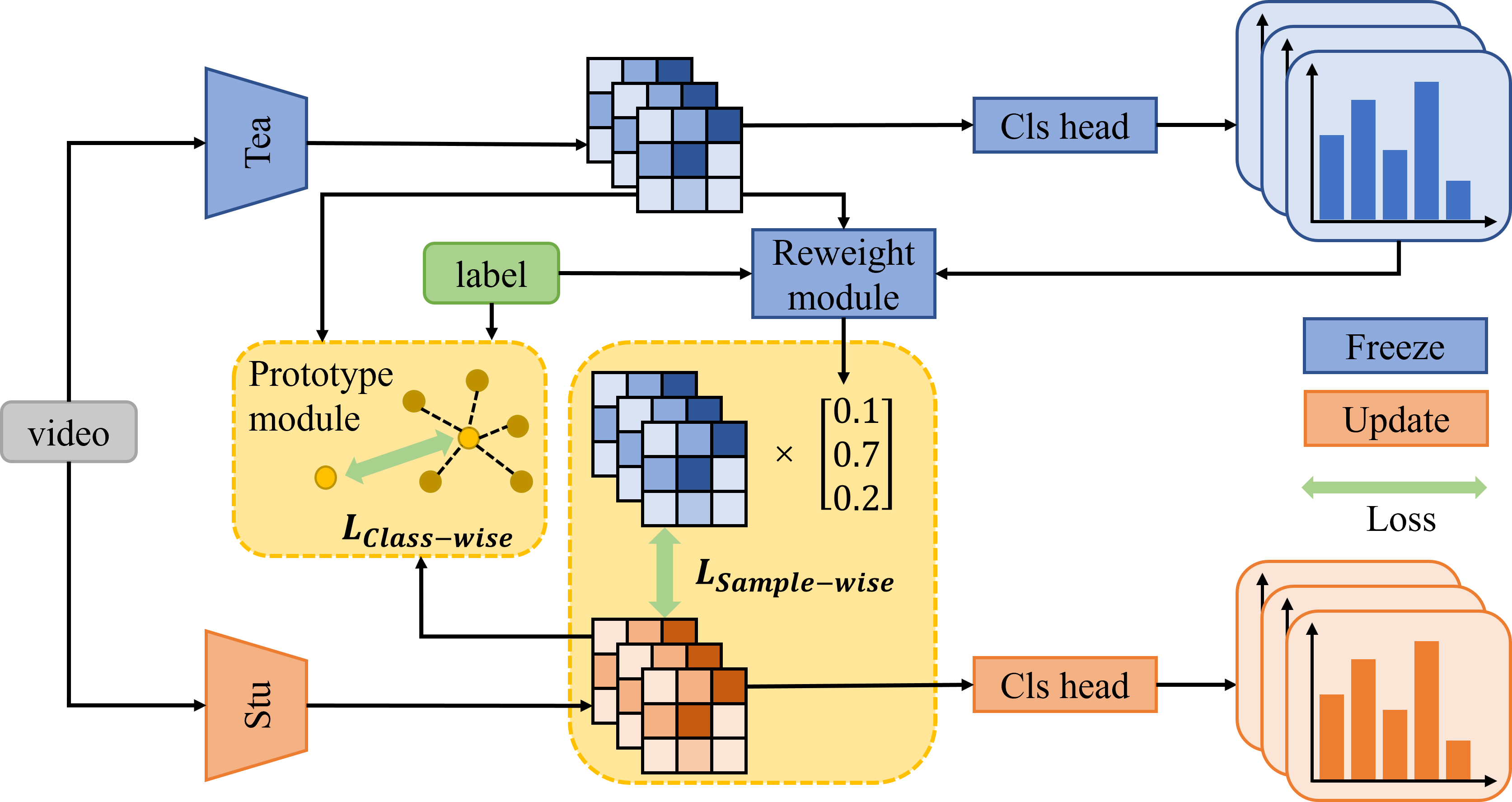}
    \caption{We propose a label-guided feature knowledge distillation framework. Our approach primarily consists of two submodules: a weight adjustment module for sample-wise distillation and a prototype network module for class-wise distillation.}
    \label{fig:method 1}
\end{figure*}

\subsection{Label-guided Knowledge Distillation}
To address the aforementioned problem, we introduce label information into traditional feature distillation to assist the student model in better learning the teacher model's feature representation. The proposed distillation framework is illustrated in Fig.~\ref{fig:method 1} and it consists of two main submodules: sample-wise knowledge distillation and class-wise knowledge distillation. In sample-wise knowledge distillation, the loss between teacher and student feature maps is weighted according to importance scores and then used as the final distillation loss. In class-wise knowledge distillation, the student model's sample features are aligned with the corresponding category feature prototypes computed by the teacher model in the feature space.
\subsubsection{Sample-wise Knowledge Distillation}
In video understanding tasks, video data is represented as a set of stacked feature maps\cite{huang2018makes}. In sample-wise knowledge distillation, we introduce a weight adjustment module to calculate the importance score of each feature map using the ground truth labels and the probability distribution produced by the teacher. This importance score is then used to adjust the weights of the loss between each pair of teacher and student feature maps, which serves as the final distillation loss for the student model.

Formally, let $q^T_i\in R^{N\times Class\times 1}$ represents the prediction result of the teacher model before Softmax function. For the video features $F^T_i$ extracted by the teacher, the corresponding feature map importance score (i.e., weight) is calculated as follows:
\begin{equation}
    {weight\_score}_i = \theta(q^T_{i,GT}),
\end{equation}
\begin{equation}
    q^T_{i,GT} = q^T_i \times GT,
\end{equation}
where, $q^T_{i,GT}\in R^{N\times 1}$ represents the probability distribution of the teacher's inference results $q^T_i$ corresponding to the ground truth label $GT$. The function $\theta(\cdot)$ is a normalization and post-processing function. In our implementation, we explore two different methods for normalization and post-processing: direct normalization and masked normalization.
\paragraph{Direct Normalization}
In direct normalization, the importance score of the feature is calculated based on the distribution of the ground truth class over the temporal dimension of the teacher's inference results. After normalization, the result is used as the importance score of the feature:
\begin{equation}
    \theta(q^T_{i,GT}) = Softmax(q^T_{i,GT}).
\end{equation}
Here, the Softmax function normalizes the importance scores to [0,1], allowing for subsequent processing.
\paragraph{Masked Normalization}
During direct normalization, the differences in the inference results across different time are reduced after normalization. Since direct normalization does not specifically account for negative values, we use masking mechanism to suppress low-importance features and highlight high-importance features. Specifically, we mask out negative results and only use positive results as valid importance scores for the subsequent distillation loss computation:
\begin{equation}
    \theta(q^T_{i,GT}) = Softmax(mask(q^T_{i,GT})) \circ \mathbb{I}(q^T_{i,GT} > 0),
\end{equation}
where $\circ$ is element-wise multiplication, $\mathbb{I}(\cdot)$ is the indicator function, the masking function $mask(\cdot)$ is defined as:
\begin{equation}
    mask(x) = \{\max (x_n, 0)\}_{n=1}^N.
\end{equation}
The final sample-wise distillation loss introduces the feature importance scores as weights based on vanilla feature distillation loss, which is defined as:
\begin{equation}
    \mathcal{L}_{Sample-wise} = \sum_{n=1}^{N} {weight\_score}_{i,n} \times \Vert f(F^T_{i,n})-f(F^S_{i,n}) \Vert_2^2,
\end{equation}
where $f(\cdot)$ is the project function.

\subsubsection{Class-wise Knowledge Distillation}
In addition to conducting one-to-one distillation of teacher-student features for the same sample through sample-wise knowledge distillation, there is also relational knowledge among features of different instances within the same category that the student model can learn\cite{yun2020regularizing}. Features of samples within the same category contain high similarity, while features of samples from different categories show low similarity. For example, consider a collection of videos labeled as "cliff diving" and "cycling". The features extracted by an excellent teacher model should form two compact feature clusters in the feature space, corresponding to two types of video samples. Aligning the features of student and teacher models at the category level can help the student model mimic high-level semantic information of the teacher model, making the feature representation of samples within the same category more compact, ultimately improving the effectiveness of knowledge distillation\cite{park2019relational}.

In this paper, we propose a class-wise knowledge distillation approach based on a prototype network. We generate corresponding feature prototypes by selecting samples correctly inferred by the teacher model according to the ground truth labels. Formally, for an action category $c$, its corresponding prototype ${prototype}_c$ is calculated as follows:
\begin{equation}
    {prototype}_c = \frac{1}{I} \times \sum_{i=1}^{I} F^T_i,
\end{equation}
where $I$ is the number of samples in the training set with ground truth label $c$, and $F^T_i$ represents the feature of sample $i$ extracted by the teacher model. The final class-wise distillation loss is defined as:
\begin{equation}
    \mathcal{L}_{Class-wise} = \sum_{c=1}^{C} \sum_{i=1}^{I} \Vert prototype_c - f(F^S_{c,i}) \Vert_2^2,
\end{equation}
where $C$ represents the number of categories, $I$ is the number of samples in each category, $f(\cdot)$ is the projection function, and $F^S_{c,i}$ denotes the features of sample $i$ with ground truth label $c$ calculated by the student model.

During training, it is impossible to obtain the features of all samples from the same category simultaneously due to memory constraints. Therefore, when calculating the category prototype, we update the prototype using a moving average method \cite{hyndman2011moving} within each batch of samples. Specifically, in each batch, we update the prototype for each category according to the following approach:
\begin{equation}
    {prototype}_c = \frac{({prototype}_c \times {proto\_cnt}_c + \sum_{i=1}^{n} F^T_{c,i})}{{proto\_cnt}_c+n},
\end{equation}
\begin{equation}
    {proto\_cnt}_c = {proto\_cnt}_c + n,
\end{equation}
where ${proto\_cnt}_c$ is the number of samples with label $c$ that have already been computed, $n$ is the number of samples with label $c$ in the current batch, and $F^T_{c,i}$ represents the features of the $i$-th sample extracted by the teacher model with label $c$. Additionally, since the teacher model does not compute the prototype features at the beginning of the training, we adopt a warm-up strategy for class-wise knowledge distillation. Specifically, during the first few training epochs, referred to the warm-up period, the teacher model updates the prototype features for each category while computing sample features, but does not perform class-wise knowledge distillation. After the warm-up period, as the prototype features are stable, at which point the teacher model stops updating the prototype features and the framework begins the class-wise knowledge distillation.

Class-wise knowledge distillation not only explicitly guides the student model to learn the relationships among sample features within the same category, but also effectively mitigates the impact of a small number of poorly extracted features by the teacher model. Specifically, when calculating the category prototypes, we use only the sample features correctly inferred by the teacher model. This approach avoids the negative impact of the teacher model's limitations on knowledge distillation. Additionally, the prototype calculation extracts consistent features inherent to samples within the same category, reducing the impact of the noise in individual samples, which has a regularization effect and ultimately improves the generalization ability of the student model\cite{park2019relational}.

\subsection{Training and Inference}
During the training process, we adopt an offline knowledge distillation approach. The teacher model's parameters are frozen after pretraining and finetuning on the training set, with only the student model's parameters being updated. Additionally, the two proposed distillation loss functions can be used either independently or in conjunction with the task-related loss function, which for action recognition tasks is the cross-entropy function. The overall training loss function can be expressed as:
\begin{equation}
    \mathcal{L} = CE(\widetilde{y},y) + \alpha \times \mathcal{L}_{Sample-wise} + \beta \times \mathcal{L}_{Class-wise},
\end{equation}
where $CE(\cdot)$ denotes the cross-entropy loss function, $\widetilde{y}$ is the predicted label by the student model, $y$ is the corresponding ground truth label, and $\alpha$ and $\beta$ are hyperparameters that balance the distillation losses. During inference, we only use the student model to predict the category label.

\section{Experiments}

In this section, we first analyze the effects of two weight adjustment module implementations. We then present the main experimental results of our proposed method, followed by ablation experiments of each module, generalization experiments, and hyperparameters sensitivity experiments.

\subsection{Experiment Setup}

\textbf{Datasets.} To validate the effectiveness of our framework, we conduct experiments on the mainstream action recognition datasets UCF101\cite{soomro2012ucf101} and HMDB51\cite{kuehne2011hmdb}.
\begin{itemize}
    \item \textbf{UCF101} consists of 13,320 real action videos across 101 action categories. It has three official data split, each dividing the data into training and testing sets with an approximately 7:3 ratio.
    \item \textbf{HMDB51} contains 6,849 videos, mostly sourced from movies, and includes 51 action categories, with each category having at least 101 video clips. It also has three official training/testing split with a 7:3 ratio. In our experiments, we selected the first split scheme for each dataset.
\end{itemize}

\textbf{Evaluation Metrics.} We follow standard evaluation protocol and use recognition accuracy as the metric for model evaluation. Specifically, we use Top-1 accuracy and Top-5 accuracy as evaluation metrics for the action recognition task.

\textbf{Implementation Details.} We use the TV-L1 algorithm\cite{brox2009large} to extract all optical flow frames for each video, with each video set to 64 frames in length. The frame size is scaled to 256 for UCF101 and 240 for HMDB51. During training, we applied data augmentation techniques including multi-scale cropping and random horizontal flipping. During inference, we only use center cropping.

We initialize the I3D model parameters with weights pretrained on ImageNet\cite{deng2009imagenet} and finetune the model for 20 epochs on the UCF101 and HMDB51 datasets to serve as the teacher model. The Top-I3D model is trained from scratch on the action recognition datasets until convergence, serving as the baseline student model.

For the baseline network on the UCF101 dataset, we use the SGD optimizer with a learning rate of ${10}^{-3}$, momentum of 0.9, and weight decay of $5\times{10}^{-4}$. The batch size is set to 12. For the HMDB51 dataset, the same SGD optimizer settings are used. For both datasets, we train the model on RGB and optical flow data for both 200 epochs. During testing, the batch size is set to 1, and the final result is obtained by averaging the predictions from the RGB and optical flow branches.

During knowledge distillation, the teacher model's parameters are frozen, and only the student model's parameters are updated. Since category prototypes are not available at the beginning of training, we adopt warm-up strategy to compute teacher prototypes, i.e., during the initial warm-up epochs, only the teacher model computes the category prototypes without performing class-wise distillation based on the prototype network. In our experiments, the warm-up period is set to 1 epoch. Note that the proposed weight adjustment module and prototype network module are used only during training, and the student model does not increase any additional parameters or computation cost during the testing phase.

\subsection{Sample-wise Distillation weight adjustment module}
We start by evaluating the effectiveness of the feature weight adjustment module. Specifically, we propose two implementation approaches for the feature weight adjustment module. The propose of this module is to use the teacher’s inference results to represent the temporal distribution of video frame features, exploiting the variation in importance along the temporal dimension.

In our implementation, we find that directly using the outputs from the teacher's logits layer achieves good results. Our two approaches show different ways to use the ground truth labels and handle negative values.

\textbf{Approach 1} uses the ground truth labels to select the distribution of the correct class in the teacher's output along the temporal dimension. After normalization and scaling, the distribution is used as the importance score for the feature maps:
\begin{equation}
    {weight\_score}_i = Softmax(q^T_i \times {GT}_i),
\end{equation}
where $q^T_i$ represents the probability distribution of the $i$-th sample generated by the teacher model, ${GT}_i$ is the ground truth label for the $i$-th sample.

\textbf{Approach 2} uses a masking mechanism to conceal results from the teacher that are negative, and only the results greater than zero are considered as valid importance scores for subsequent distillation loss calculation:
\begin{equation}
    {weight\_score}_i = Softmax(mask(q^T_i \times {GT}_i)) \circ \mathbb{I}(q^T_{i,GT} > 0),
\end{equation}
where $mask(\cdot)$ is the masking function, $\circ$ is element-wise multiplication, $\mathbb{I}(\cdot)$ is the indicator function. 

Additionally, for comparison, we aggregate the teacher's output results along the category dimension and serves as the importance score after normalization and scaling:
\begin{equation}
    {weight\_score}_i = Softmax(f_{sum}(q^T_i)),
\end{equation}
where $f_{sum}(\cdot)$ is the aggregate function. The reason for this approach is that the aggregated results represent the likelihood of an action occurring at a given moment, a higher score indicates a higher likelihood that the feature contains information about the corresponding action. Note that this method does not use label information.

\begin{table}[htbp]\centering
\caption{Comparison of different implementation methods for the weight adjustment module.}
\label{tab2}
    \setlength{\tabcolsep}{1.2mm}{
    \begin{tabular}{ccc}
    \toprule
    \textit{Method} & \textit{Top-1}(\%) & \textit{Top-5}(\%)\\
    \midrule
    Student baseline & 63.36 & 82.08\\
    Student+FeatDistill & 69.92 & 87.81\\
    Student+reweight(Approach 1) & \textbf{70.45} & \textbf{87.74}\\
    Student+mask(Approach 2) & 70.24 & 87.71\\
    Student+sum(comparison) & 70.26 & 87.55\\
    \bottomrule
\end{tabular}}
\end{table}

We conduct experiments on the UCF101 dataset for the two proposed approaches and the comparison method. We also include the accuracy of the vanilla feature distillation for comparison. The results are shown in Table~\ref{tab2}. For fairness, we use the same parameter setting for all implementations.

It can be discovered that all three approaches achieve an accuracy improvement of nearly $7\%$ compared to the baseline student model ($63.36\%$) and exceeded the vanilla feature distillation($69.92\%$). Approach 1, which provides the highest accuracy improvement, reaches a final accuracy of $70.45\%$. Approach 2 with masking mechanism and the method using probability aggregation followed in performance. Comparing approach 1 with approach 2, we conclude that directly discarding feature segments with negative scores during knowledge distillation is too aggressive. This approach also loses beneficial knowledge contained in background segments while suppressing irrelevant noise, which leads to the student model failing to capture the complete semantic features of the action. 

Comparing approach 1 with the probability aggregation method, we find that incorporating label information is crucial for approach 1's superior performance. Label information indicates the ground truth category, and adjusting features based on the correct class probability distribution is more targeted and reliable than simply aggregating probabilities across all action categories. It also highlights the importance of using ground truth information to guide feature distillation. Therefore, in subsequent experiments, we adopt approach 1 to adjust video feature weight.

\begin{table*}[htbp]\centering
\caption{Validation accuracy and computation cost on UCF101 and HMDB51. We set I3D as the teacher model, Top-I3D as the student model. For fair comparison, we keep the same training configuration for all methods.}
\label{tab3}
\setlength{\tabcolsep}{2mm}{
    \begin{tabular}{cccccccc}
    \toprule
    \multirow{2}*{Model}  &   \multirow{2}*{Knowledge} &   \multicolumn{2}{c}{UCF101}   &   \multicolumn{2}{c}{HMDB51}  &   \multirow{2}*{FLOPS(G)}    &   \multirow{2}*{Params(M)}\\
    \cmidrule(lr){3-4}\cmidrule(lr){5-6}
    & & Top-1(\%)   &   Top-5(\%)   &   Top-1(\%)   &   Top-5(\%)\\
    \midrule
    Teacher    &   -  &   90.35    &   98.76    &   69.02    &   88.76    &   111.3    &   12.7\\
    Student    &   -  &   63.36    &   82.08    &   52.35    &   74.97    &   45.5 &   10.5\\
    KD\cite{hinton2015distilling}   &   Logits  &   67.20    &   85.44    &   53.00    &   75.16    &   45.5 &   10.5\\
    CTKD\cite{li2023curriculum}   &   Logits  &   67.63   &    88.37   &    52.65   &    75.01   &   45.5 &   10.5\\
    LSKD\cite{sun2024logit}   &   Logits  &   68.33    &   86.36    &   53.20    &   75.10    &   45.5 &   10.5\\
    Feature KD\cite{romero2014fitnets}   &   Feature &   69.92   &   87.81   &   53.14   &   75.27   &   45.5    &   10.5\\
    FN\cite{xu2020feature} &   Feature  &   70.95    &   88.54    &   53.25    &   75.34    &   45.5   &   10.5\\
    MGD\cite{yang2022masked}  &   Feature &   71.43   &    89.39   &    54.15   &    76.07   &   45.5   &   10.5\\
    SimKD\cite{chen2022knowledge} &   Feature &   72.01   &    89.60   &    53.67   &    75.95   &   45.5  &   10.5\\
    AT\cite{zagoruyko2016paying} &   Attention   &   71.27    &   88.57    &   52.93    &   75.14    &   47.4    &   37.8\\
    CTKD\cite{zhao2020highlight}    &   Logits+Feature   &   71.77    &   89.14    &   53.91    &   75.87    &   45.5  &   10.5\\
    Ours   &   Feature   &   \textbf{73.51}    &   \textbf{90.30}    &   \textbf{54.77}    &   \textbf{76.41}    &   \textbf{45.5}  &   \textbf{10.5}\\
    \bottomrule
\end{tabular}}
\end{table*}

\subsection{Main Results}
To validate the effectiveness of our proposed method, we compared it with other knowledge distillation methods on the action recognition task. We used the I3D\cite{carreira2017quo} model as the teacher model and the Top-I3D\cite{xie2018rethinking} model as the student model. The Top-I3D model is derived from the I3D model by replacing some shallow 3D convolutional blocks with 2D convolutional blocks to reduce the model's parameters and computational resource requirements.

As shown in Table~\ref{tab3}, our method achieves superior improvements in student model accuracy compared to existing knowledge distillation methods on both datasets. Specifically, on the UCF101 dataset, our method improves the Top-1 accuracy of the student model by $10.15\%$ and the Top-5 accuracy by $8.22\%$, without introducing additional parameters or computational cost. Compared to our baseline algorithm, i.e., the vanilla feature distillation, our approach provides an additional $3.43\%$ and $2.20\%$ improvement in Top-1 and Top-5 accuracy, respectively. On the more challenging HMDB51 dataset, we also obtain similar results, with our method improving Top-1 and Top-5 accuracy by $2.42\%$ and $1.44\%$, respectively.

For fairness, we applied other knowledge distillation methods to the same teacher-student model pair. Given that feature-based knowledge contains more learnable information, our method easily outperforms recent state-of-the-art logits-based knowledge distillation methods. Compared to other recent feature-based knowledge distillation methods, such as MGD\cite{yang2022masked} method using a masking-recovery training paradigm and SimKD\cite{chen2022knowledge} using feature alignment and teacher classifier reuse strategy, our method also demonstrates superiority. 

We believe the improvement of our method comes from two aspects. On the one hand, the sample-wise feature distillation using the feature weight adjustment module leverages the teacher's inference results and label information as guidance. This approach allows the student model to focus on learning features based on their contribution to the correct inference result, rather than blindly learning all features extracted by the teacher. On the second hand, the class-wise feature distillation based on the prototype network further aligns the teacher and student models' category feature representations, enhancing the student model’s ability to extract high-level semantic information.

In addition to accuracy improvements, the student model shows a significant reduction in computational and parameter costs compared to the teacher model. Specifically, the student model's computational cost is reduced by $59.1\%$, and the parameters is reduced by $17.3\%$. Combined with our proposed method, this lightweight resource usage makes it feasible to deploy video analysis models on resource-constrained edge devices.

\subsection{Ablation Study}
To thoroughly investigate the validity of our proposed method, we systematically analyze the effects of each submodule and the generalization of the method across different backbone networks in this section.

\subsubsection{Contribution of Each Submodule}

\begin{table}[htbp]\centering
\caption{The impact of each submodule on model accuracy improvement on the UCF101 dataset.}
\label{tab4}
    \setlength{\tabcolsep}{1.2mm}{
    \begin{tabular}{ccccc}
    \toprule
    \textit{baseline}& $\mathcal{L}_{Sample-wise}$ & $\mathcal{L}_{Class-wise}$ & Top-1(\%) & Top-5(\%)\\
    \midrule
    $\checkmark$ & & & 63.36 & 82.08\\
    $\checkmark$  & $\checkmark$  &   &   72.64    &   89.53\\
    $\checkmark$  & &$\checkmark$  & 70.47 & 88.11\\
    $\checkmark$  & $\checkmark$    & $\checkmark$  & 73.51 & 90.30\\
    \bottomrule
\end{tabular}}
\end{table}

We first explore the impact of each submodule on the overall performance. We start with the baseline model, which is the student model trained solely with the classification loss. We then introduce sample-wise feature distillation and class-wise feature distillation separately. The former means the student model assigns different loss weights to different feature maps based on their importance, while the latter implies the student model's loss function includes the cross-entropy between the student’s inference and ground truth label, and the distance between the feature produced by the student and the teacher’s prototype feature. Finally, we incorporate both the sample-wise distillation with feature weight adjustment and the class-wise distillation with prototype network.

Table~\ref{tab4} summarizes the ablation results for these four settings on the UCF101 dataset. It can be observed that both proposed distillation methods independently improve the accuracy of the student model, and introducing two submodules together further enhances the model’s performance. Specifically, compared to the baseline model, introducing sample-wise distillation improves the student model's accuracy by $9.28\%$, indicating that our module, by adjusting feature map weights, helps the student model focus more on the teacher’s features that contain richer action information, thereby improving the knowledge distillation effect. Similarly, introducing class-wise feature distillation improves the student model's accuracy by $7.11\%$ compared to the baseline, showing that class-wise distillation also enables the student model to learn the teacher model's feature extraction capability. Finally, combining both sample-wise and class-wise distillation methods further enhances the student model’s performance, with Top-1 accuracy increasing by $10.15\%$ and Top-5 accuracy increasing by $8.22\%$ compared to the baseline, achieving a final Top-1 accuracy of $73.51\%$ and Top-5 accuracy of $90.30\%$.

\subsubsection{Effectiveness of Weight Adjustment Module}

\begin{table}[htbp]\centering
\caption{Comparison between the effects of sample-wise knowledge distillation and traditional feature distillation on the UCF101 dataset.}
\label{tab5}
    \setlength{\tabcolsep}{1.2mm}{
    \begin{tabular}{ccccc}
    \toprule
    \textit{baseline}&Feature Dist. & Sample-wise Dist. & Top-1(\%) & Top-5(\%)\\
    \midrule
    $\checkmark$ & & & 63.36 & 82.08\\
    $\checkmark$  & $\checkmark$  &   &   69.92    &   87.81\\
    $\checkmark$  & & $\checkmark$  & \textbf{72.64} & \textbf{89.53}\\
    \bottomrule
\end{tabular}}
\end{table}

Our proposed weight adjustment module for sample-wise distillation is an improvement over vanilla feature distillation methods. To demonstrate that the effectiveness of the weight adjustment module is not solely due to vanilla feature distillation, we compare the results of sample-wise distillation with those of vanilla feature distillation. Experiments are conducted on the UCF101 dataset, and the results are shown in Table~\ref{tab5}.

As observed, our proposed sample-wise distillation method surpasses vanilla feature distillation method by nearly $2.72\%$ in Top-1 accuracy. This indicates that the accuracy improvement provided by our sample-wise distillation method is not only due to vanilla feature distillation but also due to its ability to selectively focus on frames that are more critical for correct inference.
\subsection{Generalization of Our Method}

\begin{table}[tbp]\centering
\caption{The improvement of our proposed method on different student models on the UCF101 dataset.}
\label{tab6}
    \setlength{\tabcolsep}{5mm}{
    \begin{tabular}{ccc}
     \toprule
     Model         & Top-1(\%) & Top-5(\%)\\
     \midrule
     Top-I3D       & 63.36  & 82.08\\
     Top-I3D+Ours    & \textbf{73.51}  & \textbf{90.30}\\
     Bottom-I3D    & 53.64  & 72.69\\
     Bottom-I3D+Ours & \textbf{57.73}  & \textbf{76.87}\\
     I2D           & 56.07  & 77.24\\
     I2D+Ours        & \textbf{64.98}  & \textbf{83.27}\\
     \bottomrule
\end{tabular}}
\end{table}

To demonstrate the generalization of our proposed method, we select a series of variant I3D models\cite{xie2018rethinking} as student models. Specifically, in addition to the Top-I3D model, we also chose Bottom-I3D and I2D models for experiments. The results are shown in Table~\ref{tab6}.

As observed, for Top-I3D model, our method improve Top-1 and Top-5 accuracy by $10.15\%$ and $8.22\%$, respectively, achieving final accuracies of $73.51\%$ and $90.30\%$. For Bottom-I3D model, which is derived by replacing deep 3D convolutional layers with 2D convolutional layers, our method improve Top-1 and Top-5 accuracy by $4.09\%$ and $4.18\%$, respectively, reaching final accuracies of $57.73\%$ and $76.87\%$. Lastly, for I2D model, which is composed entirely of 2D convolutional layers, our method improve Top-1 and Top-5 accuracy by $8.91\%$ and $6.03\%$, respectively, achieving final accuracy of $64.98\%$ and $83.27\%$.

These results demonstrate that our knowledge distillation method is effective even when transferring knowledge from a pure 3D network teacher to a pure 2D network student, proving the method's effectiveness across different network architectures.

\subsection{Hyperparameter Sensitivity Study}
In our method, the hyperparameters $\alpha$ and $\beta$ serve as the weights for the loss functions in sample-wise distillation and class-wise distillation, respectively. Table~\ref{tab7} and Table~\ref{tab8} show the impact of these hyperparameters on the method's performance improvement on the UCF101 dataset when set to different values. Notably, since the method showed the most significant improvement when $\alpha=4$, we retained $\alpha=4$ in the sensitivity experiments for $\beta$.

\begin{table}[htbp]\centering
\caption{Sensitivity experiment of the hyperparameter $\alpha$ on the UCF101 dataset}
\label{tab7}
\setlength{\tabcolsep}{0.5mm}{
    \begin{tabular}{ccccccccc}
    \toprule
     \multirow{2}*{Parameter}  &   \multicolumn{2}{c}{1} &   \multicolumn{2}{c}{2}   &   \multicolumn{2}{c}{4}  &   \multicolumn{2}{c}{8}\\
     \cmidrule(lr){2-3}\cmidrule(lr){4-5}\cmidrule(lr){6-7}\cmidrule(lr){8-9}
      & Top-1   &   Top-5   &   Top-1   &   Top-5&   Top-1   &   Top-5&   Top-1   &   Top-5\\
     \midrule
     $\alpha$ &   70.92   &   88.50   &   71.56   &   88.98   &   \textbf{72.64}   &   \textbf{89.53}   &   72.32   &   89.85\\
     \bottomrule
\end{tabular}}
\end{table}

\begin{table}[htbp]\centering
\caption{Sensitivity experiment of the hyperparameter $\beta$ on the UCF101 dataset}
\label{tab8}
\setlength{\tabcolsep}{0.5mm}{
    \begin{tabular}{ccccccccc}
    \toprule
     \multirow{2}*{Parameter}  &   \multicolumn{2}{c}{1} &   \multicolumn{2}{c}{2}   &   \multicolumn{2}{c}{4}  &   \multicolumn{2}{c}{8}\\
     \cmidrule(lr){2-3}\cmidrule(lr){4-5}\cmidrule(lr){6-7}\cmidrule(lr){8-9}
      & Top-1   &   Top-5   &   Top-1   &   Top-5&   Top-1   &   Top-5&   Top-1   &   Top-5\\
     \midrule
     $\beta$ &   73.38   &   90.06   &   \textbf{73.51}   &   \textbf{90.30}   &   73.35   &   90.01   &   72.91   &   89.88\\
     \bottomrule
\end{tabular}}
\end{table}

As shown in tables, regardless of the values of $\alpha$ and $\beta$, the model's accuracy always improves compared to the baseline accuracy. Specifically, when $\alpha=4$ and $\beta=2$, the method achieves the most significant accuracy improvement, reaching a Top-1 accuracy of $73.51\%$ and a Top-5 accuracy of $90.30\%$. Therefore, we keep these hyperparameters throughout the experiments.

This sensitivity analysis confirms that our method is robust across different hyperparameter settings, with the chosen values of $\alpha=4$ and $\beta=2$ providing the optimal balance for performance improvement.

\section{Conclusion}

In this paper, we propose Label-Guided Knowledge Distillation to explicitly exploit the temporal importance of teacher features. Our approach conducts knowledge distillation at both the sample-wise and class-wise levels. By introducing sample-wise knowledge distillation, the student model can selectively learn teacher features, focusing more on those that significantly impact classification results. By introducing class-wise knowledge distillation, the student model can acquire relational knowledge among samples within the same class, mimicking the teacher's ability to capture high-level semantic information, thereby compacting feature representations and improving model generalization. Experimental results show that our method effectively enhances the capabilities of lightweight 3D-CNN models, further exploring the trade-off between resource efficiency and model accuracy. Additionally, this paper provides insightful research into the full utilization of label information in knowledge distillation. For future work, we believe that exploring finer-grained label information to guide feature distillation is a promising direction.

\end{document}